\documentclass[sigconf]{acmart}
\usepackage{multirow}
\usepackage{booktabs}
\usepackage{tabularx}
\usepackage{threeparttable}
\usepackage{makecell}
\usepackage{array}
\usepackage{adjustbox}
\usepackage{rotating}
\usepackage{lscape}
\usepackage{graphicx}
\AtBeginDocument{%
  }

\setcopyright{acmlicensed}
\copyrightyear{2018}
\acmYear{2018}
\acmDOI{XXXXXXX.XXXXXXX}
\acmConference[KDD]{Make sure to enter the correct
  conference title from your rights confirmation email}{August 01--05,
  2027}{San Jose}

\acmISBN{978-1-4503-XXXX-X/2018/06}

\begin{document}
\settopmatter{printacmref=false}
\setcopyright{none}
\renewcommand\footnotetextcopyrightpermission[1]{}
%%
%% The "title" command has an optional parameter,
%% allowing the author to define a "short title" to be used in page headers.
\title{Chem World: A Large-Scale Benchmark and Physics-Informed Framework for Trustworthy Chemical Property Prediction}

%%
%% The "author" command and its associated commands are used to define
%% the authors and their affiliations.
%% Of note is the shared affiliation of the first two authors, and the
%% "authornote" and "authornotemark" commands
%% used to denote shared contribution to the research.
%% 紧凑型作者与机构列表 (CVPR 风格)
\author{
  Tianyou Bai$^{1,2}$, 
  Huan Wang$^{1}$, 
  Mingchen Gao$^{2}$, 
  Fangyue Lin$^{1,3}$, 
  Pinze Ren$^{1,4}$, 
  Zhenlin Zhao$^{1}$, 
  Siming Dong$^{1}$
}

\affiliation{%
  \institution{$^1$Cleer Science}
  \country{}
}
\affiliation{%
  \institution{$^2$Institute of Automation, Chinese Academy of Sciences}
  \country{}
}
\affiliation{%
  \institution{$^3$Department of Chemistry, Imperial College London}
  \country{}
}
\affiliation{%
  \institution{$^4$Department of Chemical Engineering, Tsinghua University}
  \country{}
}

%% 邮箱统一放在页脚或紧跟作者（可选）
\email{baitianyou2024@ia.ac.cn, wh.2021@tsinghua.org.cn, {zhenlinzhao, william.d}@cleerlab.com}

\affiliation{%
  \institution{%
    \vspace{0.5em} % 稍微调整与上方文字的距离
    \includegraphics[height=1.1cm]{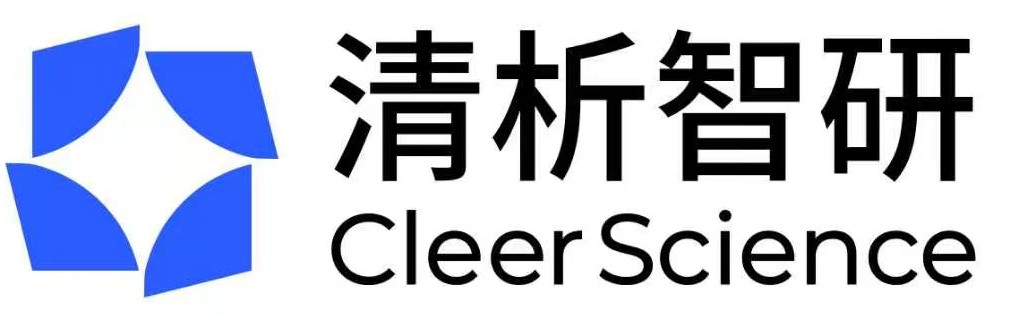} \quad % 替换为你的公司 Logo 文件名
    \includegraphics[height=1.2cm]{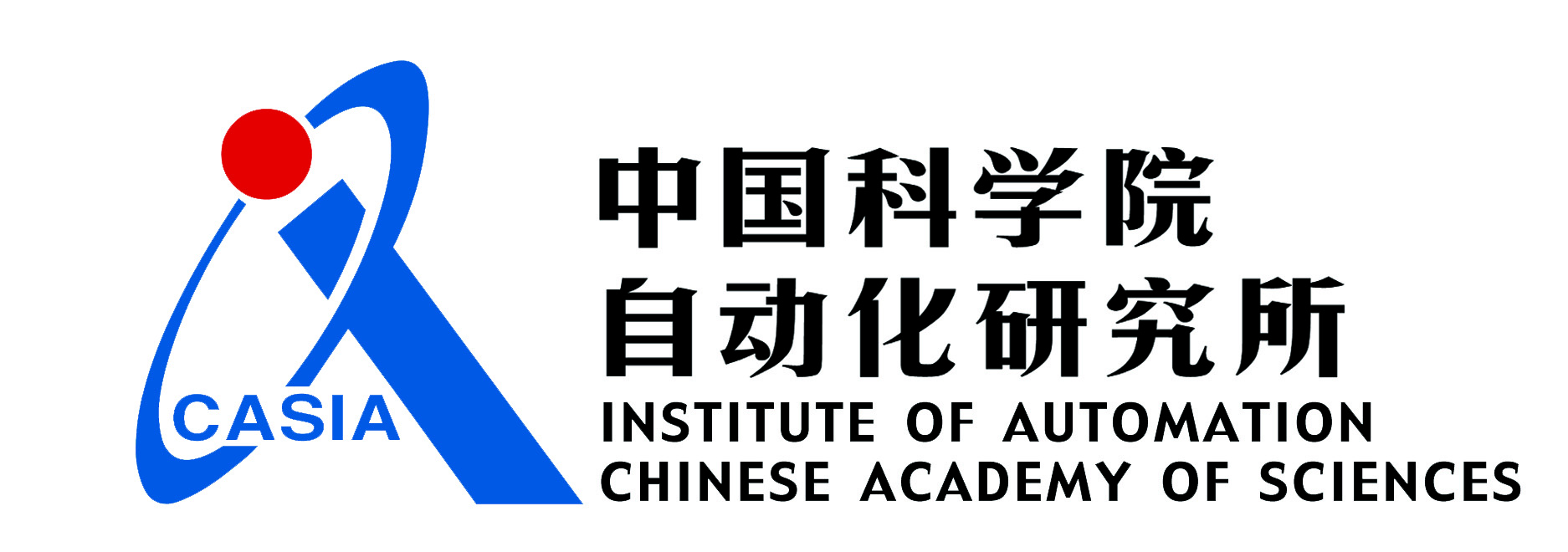} \quad   % 替换为你的学校 Logo 文件名
    \includegraphics[height=1.1cm]{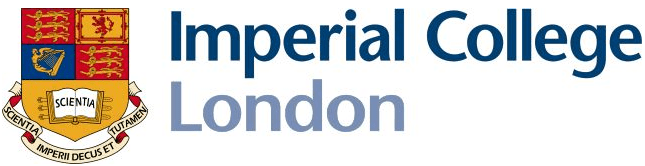} \quad 
    \includegraphics[height=1.2cm]{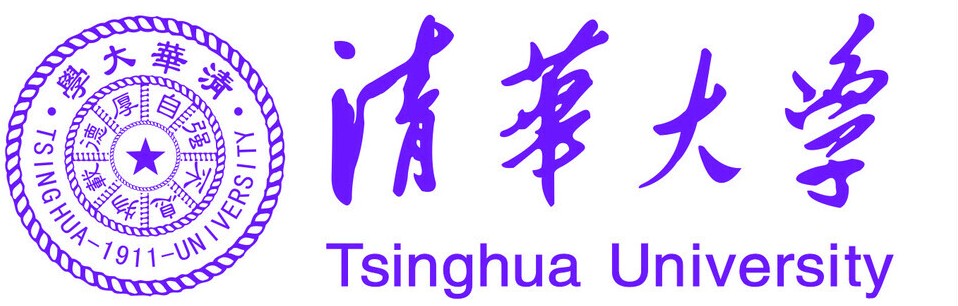}    % 替换为其他 Logo
    % 替换为其他 Logo
  }%
  \country{}
}

%%
%% By default, the full list of authors will be used in the page
%% headers. Often, this list is too long, and will overlap
%% other information printed in the page headers. This command allows
%% the author to define a more concise list
%% of authors' names for this purpose.
\renewcommand{\shortauthors}{Bai et al.}

%%
%% The abstract is a short summary of the work to be presented in the
%% article.
\begin{abstract}
Chemical property prediction plays a critical role in accelerating scientific discovery in chemistry, materials science, and drug development. However, existing benchmarks often suffer from limited task diversity, fragmented datasets, and inconsistent evaluation protocols, making it challenging to systematically assess the reliability and generalization of AI models. In this work, we introduce Chem World, a comprehensive benchmark for chemical property prediction that integrates 17 diverse chemical datasets with over 800,000 molecular samples, covering various properties including density, electrical conductivity, solubility, and other molecular characteristics. Chem World provides a unified platform for evaluating AI models across multiple property prediction tasks. Furthermore, we propose Mixture-PINN, a physics-informed neural network based prediction framework that incorporates chemical prior knowledge into data-driven learning, improving the accuracy, robustness, and reliability of chemical property prediction. Extensive experiments on Chem World demonstrate the effectiveness of our approach compared with existing methods. By combining large-scale standardized evaluation with physics-informed learning, Chem World establishes a foundation for developing trustworthy AI systems for computational chemistry and advancing AI-driven scientific discovery.
\end{abstract}

%%
%% The code below is generated by the tool at http://dl.acm.org/ccs.cfm.
%% Please copy and paste the code instead of the example below.
%%


%%
%% Keywords. The author(s) should pick words that accurately describe
%% the work being presented. Separate the keywords with commas.
%% A "teaser" image appears between the author and affiliation
%% information and the body of the document, and typically spans the
%% page.

% \received{20 February 2007}
% \received[revised]{12 March 2009}
% \received[accepted]{5 June 2009}

%%
%% This command processes the author and affiliation and title
%% information and builds the first part of the formatted document.
\maketitle

\section{Introduction}
Accurate prediction of chemical properties is fundamental to numerous scientific and industrial applications, including materials discovery, chemical engineering, and pharmaceutical development. Reliable estimation of properties such as density, electrical conductivity, solubility, and other physicochemical characteristics can significantly reduce the cost of experimental trials and accelerate the discovery of novel chemicals and materials. With the rapid development of artificial intelligence (AI), data-driven approaches have emerged as powerful tools for chemical property prediction by learning complex relationships between molecular structures and target properties.

Recent advances \cite{shi2026multi,amiri2026physics,jain2013commentary} in graph neural networks (GNNs), transformer-based molecular models, and large-scale chemical representation learning have substantially improved the performance of AI-driven chemical property prediction. However, most existing approaches primarily rely on statistical correlations extracted from data, while overlooking the underlying physical and chemical principles governing molecular behaviors. This limitation may lead to physically inconsistent predictions, poor generalization under limited data conditions, and reduced reliability when applied to complex chemical systems, particularly for mixture-based property prediction scenarios. Therefore, developing trustworthy AI models that integrate domain knowledge with data-driven learning remains a critical challenge for computational chemistry.

Meanwhile, existing benchmarks \cite{rajaonson2026chemixhub,kavian2026large} for chemical property prediction still face several limitations. Many widely used datasets focus on specific molecular tasks or individual properties, resulting in fragmented evaluation protocols and limited coverage of diverse chemical systems. The lack of a unified and comprehensive benchmark makes it difficult to systematically compare different AI models and evaluate their generalization ability across various chemical properties. A large-scale benchmark that integrates diverse datasets and provides standardized evaluation is therefore essential for advancing AI-driven chemical discovery.

\begin{figure}
    
    \centering
    \includegraphics[width=0.95\linewidth]{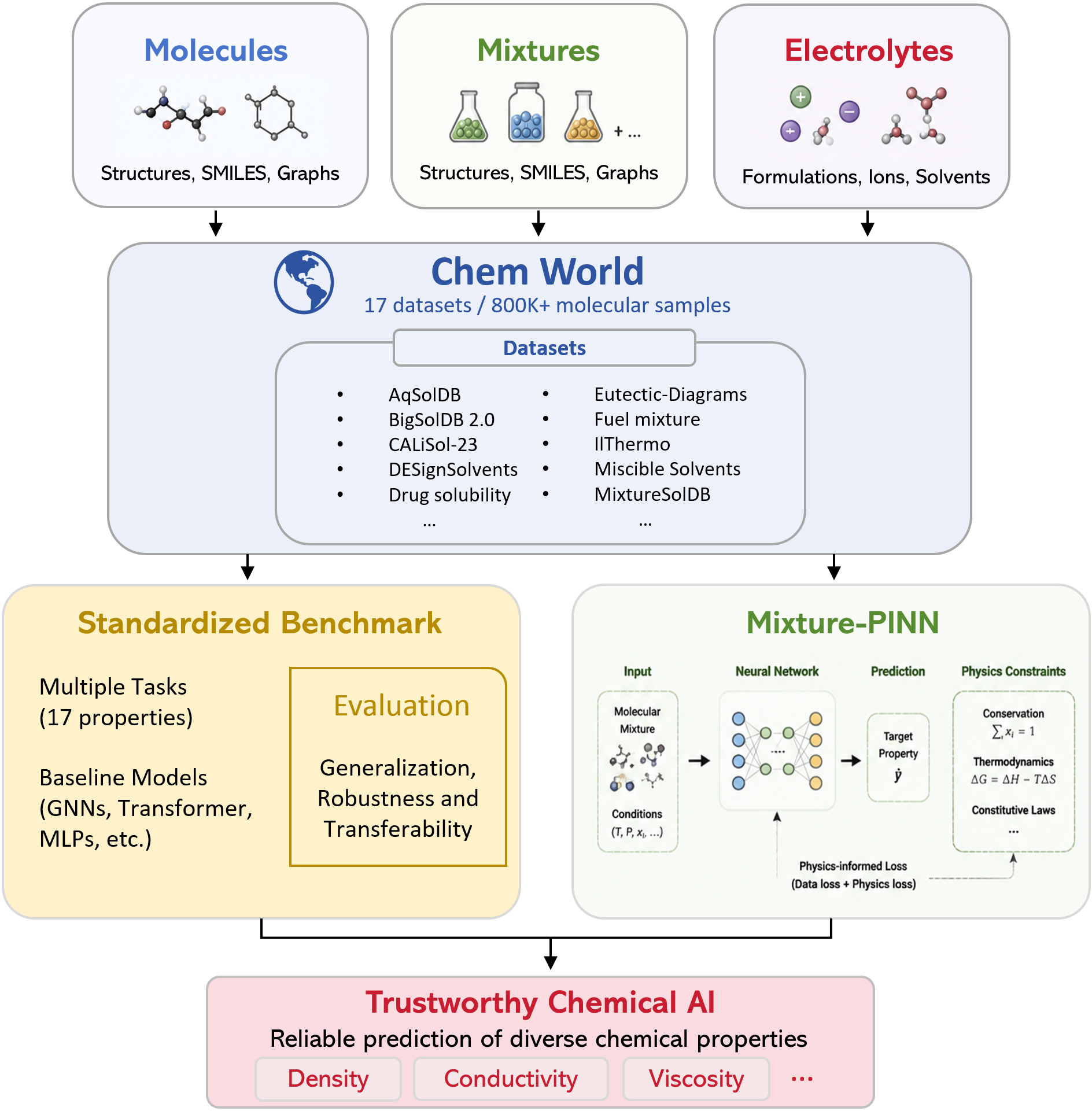}
    \captionsetup{belowskip=-18pt}
    \caption{Pipeline of Chem World and Mixture-PINN. Chem World provides standardized multi-source chemical data for comprehensive model evaluation, while Mixture-PINN integrates physical constraints to improve mixture property prediction.}
    \label{fig:1}
\end{figure}

To address these challenges, we introduce Chem World, a comprehensive benchmark for chemical property prediction that integrates 17 diverse chemical datasets \cite{sorkun2019aqsoldb,krasnov2025bigsoldb,de2024calisol,odegova2024designsolvents,bao2024towards,kumar2025electrolytomics,fajar2024estimating,kuzhagaliyeva2022artificial,chew2025leveraging,malikov2026dataset,zhang2023molsets,kazakov2012nist,tom2025molecules,bradford2023chemistry} with over 800,000 molecular instances. As shown in Fig.\ref{fig:1}, Chem World covers a broad range of physicochemical properties, including density, conductivity, solubility, and other chemical characteristics, providing a unified platform for evaluating AI models across multiple prediction tasks. Beyond benchmark construction, we further propose Mixture-PINN, a physics-informed neural network framework designed for reliable chemical property prediction. By incorporating physical and chemical prior knowledge into the learning process, Mixture-PINN enhances the consistency between model predictions and underlying chemical principles, improving prediction accuracy, robustness, and generalization ability.

Extensive experiments conducted on Chem World demonstrate that Mixture-PINN achieves competitive performance compared with existing state-of-the-art approaches across diverse chemical property prediction tasks. Our work establishes a scalable benchmark and a physics-informed learning framework toward trustworthy AI for chemistry, providing new opportunities for efficient and reliable scientific discovery.

The main contributions of this work are summarized as follows:
\begin{itemize}
\item We introduce Chem World, a large-scale benchmark for chemical property prediction that unifies 17 datasets containing over 800K molecular instances and enables standardized evaluation across diverse physicochemical properties.
\item We propose Mixture-PINN, a physics-informed neural network framework that integrates chemical knowledge into data-driven prediction models to improve the reliability and generalization of chemical property prediction.
\item We conduct extensive experiments on Chem World and demonstrate the effectiveness of physics-informed learning for developing trustworthy AI models in computational chemistry.
\end{itemize}
\section{Related Work}
\textbf{Chemical property prediction benchmarks.}
Molecular property benchmarks span several chemical domains. Quantum chemistry datasets \cite{ramakrishnan2014quantum} contain about 130K small organic molecules and 19 DFT-calculated properties for evaluating molecular representations. Benchmark suites \cite{wu2018moleculenet} cover aqueous solubility, hydration free energy, lipophilicity, and toxicity, while large-scale datasets \cite{hu2020open} provide millions of molecules for graph regression. Specialized resources address protein-ligand interactions \cite{wang2004pdbbind} and inorganic materials \cite{jain2013commentary}. For chemical mixtures, the open NIST thermophysical archive provides structured data for temperature-dependent properties \cite{kazakov2012nist}, while DETHERM \cite{westhaus1999detherm} and the Dortmund Data Bank \cite{onken1989dortmund} collect broader thermophysical measurements but require commercial access. Much of the remaining evidence is dispersed across studies of binary systems, including phase equilibria, density, surface tension, viscosity \cite{uceda2025experimental}, mixing enthalpy \cite{podgorsek2016mixing}, and excess properties \cite{li2025thermophysical,zang2025excess}. Fragmented sources, inconsistent evaluation protocols, and limited coverage of complex multicomponent mixtures hinder reliable comparison, motivating a unified and comprehensive benchmark for trustworthy chemical AI systems.

\textbf{Deep learning for molecular property prediction.}
Deep learning learns molecular representations from graphs, coordinates, and chemical strings. Chemistry-specific GNNs include formulation graphs for battery-electrolyte performance \cite{sharma2023formulation}, chemistry-aware GNNs for molecular mixtures \cite{zhang2024learning}, and geometric deep learning for electrolyte prediction and optimization \cite{zhu2024differentiable}. For three-dimensional properties, SchNet \cite{schutt2017schnet}, DimeNet \cite{gasteiger2020directional}, PaiNN \cite{schutt2021equivariant}, GemNet \cite{gasteiger2020fast}, and NequIP \cite{batzner20223} encode atomic coordinates, directional interactions, or physical symmetries to predict energies and forces. Graphormer \cite{ying2021transformers} instead uses graph-aware transformer attention, while ChemBERTa \cite{chithrananda2020chemberta} pretrains on SMILES. These models achieve strong accuracy but often lack explicit chemical priors and physical constraints.

\textbf{Physics-informed learning for chemical property prediction.}
PINNs incorporate governing equations through residual, boundary, and initial-condition losses \cite{raissi2019physics}. Chemistry-oriented methods \cite{rittig2023gibbs,ihunde2022application,wahyudi2025generalizable} impose physical structure more directly: sGDML encodes energy conservation and molecular symmetries \cite{chmiela2018towards}, PhysNet predicts energies, forces, dipoles, and charges with electrostatic interactions \cite{unke2019physnet}, and NewtonNet embeds Newtonian force relations and rotational equivariance \cite{haghighatlari2022newtonnet}. Group-contribution PINNs combine physical correlations with neural learning for thermophysical properties \cite{babaei2023physics}, while physics-informed neural operators learn solution mappings across families of differential equations \cite{li2024physics}. Together, these methods use constrained losses, conserved quantities, equivariant architectures, or physics-derived features, but standardized chemical benchmarks rarely compare them across diverse properties.

\begin{figure*}
    \centering
    \includegraphics[width=0.9\linewidth]{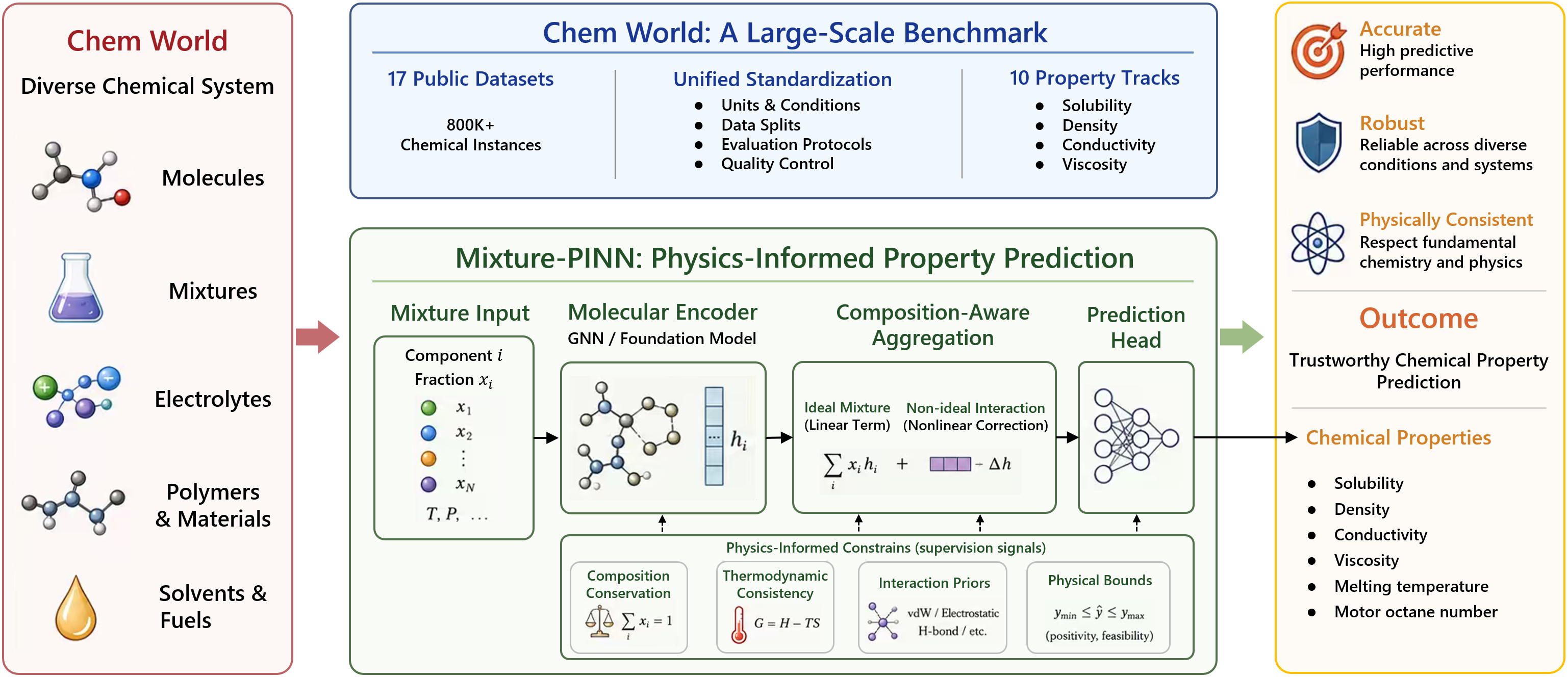}
    \captionsetup{belowskip=-10pt}
    \caption{Illustration of the Chem World benchmark and Mixture-PINN. Chem World provides a standardized large-scale testbed for diverse chemical systems. Mixture-PINN incorporates composition-aware aggregation and physics-informed constraints to predict mixture chemical properties while respecting fundamental physical and chemical laws.}
    \label{fig:2}
\end{figure*}

\section{Chem World}
Existing chemical property prediction benchmarks have primarily focused
on individual molecules, where models learn mappings between molecular
structures and their associated properties. However, many real-world
chemical systems are mixtures composed of multiple interacting
components, such as electrolytes, solvents, liquid formulations, and
eutectic systems. The emergent properties of these systems cannot be
fully determined by individual molecular structures, but depend on
component composition, intermolecular interactions, and environmental
conditions.

Although several datasets have been developed for specific mixture
properties, existing resources remain highly fragmented, with different
data formats, property definitions, and evaluation protocols. This
fragmentation prevents systematic comparison of machine learning models
and limits the development of general-purpose mixture intelligence
models.

To address this challenge, we introduce Chem World, a large-scale and
multi-property benchmark for chemical property prediction. As shown in Fig.\ref{fig:2},
unlike existing mixture datasets that focus on individual
properties or specific chemical domains, Chem World provides a unified
benchmarking framework covering diverse chemical systems, property
categories, and evaluation settings.  Beyond aggregating existing data,
Chem World establishes standardized molecular representations, mixture
formulations, task definitions, and generalization protocols to support
the development of scalable and physically meaningful AI models for
chemical mixtures.

\subsection{Motivation and Design Principles}

Chem World is designed according to three principles: diversity,
unification, and generalization.

\textbf{Chemical diversity.}
Real chemical applications involve heterogeneous mixture systems rather
than isolated molecules. Therefore, Chem World incorporates multiple
types of chemical mixtures, including solvent mixtures, electrolytes,
deep eutectic solvents, aqueous solutions, and formulation systems.
This diversity enables models to learn general mixture representations
rather than property-specific correlations.

\textbf{Unified prediction protocol.}
Existing datasets often differ in molecular representations,
composition formats, target definitions, and evaluation metrics,
making direct comparison between models difficult. Chem World provides
a unified data schema that represents each mixture using molecular
components, their corresponding fractions, optional experimental
conditions, and target properties. All prediction tasks follow
consistent preprocessing, splitting, and evaluation protocols.

\textbf{Generalizable mixture intelligence.}
Beyond conventional random train-test evaluation, Chem World is designed
to investigate whether models can generalize to unseen chemical
combinations. Therefore, the benchmark supports challenging evaluation
settings, including unseen molecular components and composition
distribution shifts.

\subsection{Benchmark Construction}

The construction of Chem World consists of four major stages:
data collection, normalization, quality control, and task organization.

\textbf{Data collection.}
We collect publicly available chemical mixture datasets from different
domains, including solubility, electrolyte systems, solvent mixtures,
thermodynamic measurements, and formulation databases. In total,
Chem World integrates 17 representative datasets containing more than
800K mixture-property records.

\textbf{Data normalization.}
Since different datasets use heterogeneous representations, we convert
all molecular components into canonical molecular formats and unify
mixture representations. Each sample is represented as \(\mathcal{M}=\{(m_i,x_i)\}_{i=1}^{N}\), where $m_i$ denotes the molecular representation of the $i$-th component,
$x_i$ represents its fraction, and $N$ is the number of components in
the mixture.

All target properties are converted into standardized numerical values
with consistent units whenever possible.

\textbf{Quality control.}
To ensure benchmark reliability, we remove duplicated records,
inconsistent molecular structures, invalid compositions, and samples
with missing target values. Furthermore, chemically equivalent
molecules are standardized using canonicalization procedures.

\textbf{Task organization.}
After preprocessing, samples are organized into nine major prediction
tracks covering thermodynamic, transport, electrochemical, and
mixture-specific properties.

\subsection{Multi-scale Prediction Tracks}

Chem World organizes chemical mixture prediction into three levels
according to the complexity of chemical interactions.

\textbf{Molecular-component level.}
This level focuses on properties primarily related to molecular
structures, where pretrained molecular encoders can provide informative
representations.

\textbf{Mixture-composition level.}
At this level, models must aggregate multiple molecular representations
according to mixture composition. Properties such as density,
viscosity, and solubility require modeling the collective behavior
of multiple components.

\textbf{Interaction-emergent level.}
Certain properties emerge from complex intermolecular interactions,
including conductivity, enthalpy of mixing, and eutectic behavior, which require models to capture non-additive effects beyond
simple component averaging.

This multi-scale organization enables Chem World to evaluate whether AI
models can progressively move from molecular understanding toward
general chemical mixture intelligence. Based on this organization, we summarized and merged numerous datasets and ultimately formed 10 property prediction tracks, including conductivity, density, enthalpy of mixing, fusion enthalpy, heat of vaporization, melting temperature, motor octane number, olfactory perceptual similarity, solubility, and viscosity prediction.

\begin{figure}
    
    \centering
    \includegraphics[width=0.95\linewidth]{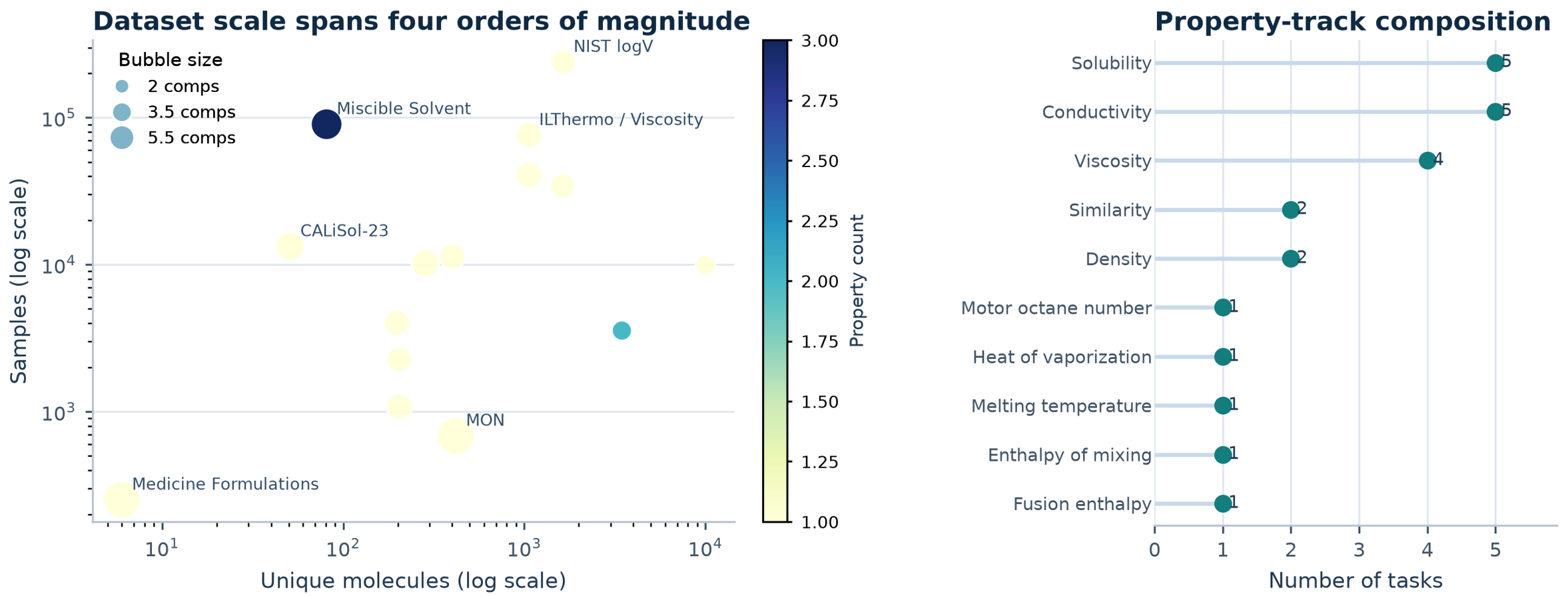}
    \captionsetup{belowskip=-24pt}
    \caption{Dataset statistics and structural composition of Chem World.}
    \label{fig:3}
\end{figure}

\subsection{Dataset Statistics and Comparison}
To provide a comprehensive overview of the Chem World benchmark, we analyze its structural distribution and property composition in Figure \ref{fig:3}. As shown in Fig.\ref{fig:3} (Left), the sub-datasets in Chem World span over four orders of magnitude in both sample size ($10^2$ to $10^5$) and molecular diversity ($10^1$ to $10^4$ unique molecules). Unlike conventional single-molecule datasets, Chem World explicitly incorporates multi-component systems, ranging from binary mixtures to complex multi-component formulations (indicated by bubble sizes up to 5.5 components). Furthermore, individual datasets cover varied property densities (indicated by color gradient), accommodating both single-property and multi-property evaluation regimes. 

As shown in Fig.\ref{fig:3} (Right), Chem World systematically organizes these sub-datasets across 10 distinct chemical property tracks, comprising a total of multiple downstream tasks. Core transport and thermodynamic properties, such as solubility, conductivity, and viscosity, feature higher task numbers (up to 5 tasks per track) to facilitate robust multi-task learning and dynamic cross-domain validation, whereas specialized thermal and physical phenomena provide targeted benchmarks for focused modeling. 

Table~\ref{tab:benchmark_comparison} compares Chem World with existing
mixture benchmarks. Different from previous resources, Chem World provides broader property
coverage, larger scale, and unified evaluation protocols. In particular,
Chem World includes both conventional physicochemical properties and
emergent mixture behaviors, enabling comprehensive evaluation of mixture
representation learning.
\begin{table}[t]
\centering
\caption{
Comparison between Chem World and existing chemical mixture benchmarks.
}
\label{tab:benchmark_comparison}
\resizebox{\linewidth}{!}{
\begin{tabular}{lccccc}
\toprule
Benchmark
& Scale
& Properties
& Mixture Types

& OOD Evaluation
\\
\midrule

CheMixHub
& $\sim$ 500$\text{K}$
& Multiple
& Limited
& Yes
\\

MixtureSolDB
& 175K+
& Solubility
& Solvent mixtures
& No
\\

MolSets
& 1K+
& Multiple
& Molecular sets
& No
\\

Chem World
&800K+
&10+
& Diverse
&Yes
\\

\bottomrule
\end{tabular}
}
\vspace{-20pt}
\end{table}

\section{Mixture-PINN}
\subsection{Model Overview}
Chemical mixtures differ from individual molecules because their
properties emerge from collective behaviors among multiple components.
A successful mixture model should therefore satisfy three requirements:
preserving molecular-level chemical information,
capturing nonlinear interactions among components, and
maintaining physically meaningful representations.

To address these challenges, we propose Mixture-PINN, a
physics-grounded mixture representation framework that integrates
chemical priors into neural representation learning. Given a mixture consisting of \(N\) molecular components \(\mathcal{M}=\{(m_i,x_i)\}^N_{i=1}\), the objective of Mixture-PINN is to learn a mapping: \(f_\theta=(\mathcal{M},T,P)\rightarrow y\), where \(y\) denotes the target mixture property, and \(T\), \(P\) represent environmental conditions such as temperature and pressure. 

Different from conventional PINNs that solve physical differential
equations, Mixture-PINN incorporates differentiable physical constraints
into the learning process to regularize mixture representations.
The framework consists of three components: a molecular encoder that extracts component-level representations, an interaction-aware mixture aggregator that models nonlinear
relationships among components, and a physics-guided learning module that introduces composition,
interaction, and property-specific consistency constraints.

\subsection{Molecular Representation}
For each module \(m_i\), we first obtain its latent representation through a pretrained molecular encoder: \(h_i=E_\phi(m_i)\), where \(E_\phi\) can be implemented using molecular foundation models, such as MolT5 or MolFormer.

However, molecular mixtures cannot be represented by simply averaging individual molecular embeddings, because mixture properties are highly dependent on component ratios and intermolecular interactions. Therefore, we introduce an interaction-aware aggregation module. For each molecular pair $(i,j)$, an interaction representation is
computed as \(e_{ij}=\phi(h_i,h_j)\), where $\phi(\cdot)$ is a lightweight neural network.

The final mixture representation is obtained by combining individual
component information and pairwise interaction information, as shown in Eq.\ref{eq:1},
\begin{equation}
  z_m=
\sum_i\alpha_i h_i+
\sum_{i,j}\beta_{ij}e_{ij},
  \label{eq:1}
\end{equation}
where $\alpha_i$ and $\beta_{ij}$ are learnable attention weights. This design enables Mixture-PINN to capture both additive contributions
and non-additive interaction effects.

% \begin{equation}
%   \Delta h=\Psi_\theta(h_1,h_2,\dots,h_N,x_1,x_2,\dots,x_N),
%   \label{eq:1}
% \end{equation}
% where \(\Psi_\theta\) is a neural interaction network modeling non-ideal mixing effects. The final mixture representation is shown in Eq.\ref{eq:2}.
% \begin{equation}
%   z_{mix}=\sum_ix_ih_i+\Delta h,
%   \label{eq:2}
% \end{equation}
% where the first term provides physically meaningful mixture priors, while the second term captures complex molecular interactions beyond classical mixing rules.
\subsection{Physics-Informed Constraint Learning}
Although deep neural networks can effectively learn complex mappings
from molecular representations to mixture properties, purely
data-driven approaches may capture statistical correlations that violate
basic chemical principles. To improve the physical consistency of learned
representations, Mixture-PINN introduces differentiable
physics-grounded constraints during training.

Different from conventional physics-informed neural networks that
require explicit physical equations or simulation data, our framework
incorporates general chemical priors as regularization terms. These
constraints do not require additional physical annotations and can be
adapted to heterogeneous chemical mixture datasets.

The contribution of each component in a mixture is closely related to
its composition fraction. However, directly enforcing
$\sum_i x_i=1$ provides no effective learning signal because mixture
fractions are already normalized during preprocessing. Therefore, we
instead align the learned contribution weights with the observed
composition information.

Specifically, the composition consistency loss is defined as shown in Eq.\ref{eq:2},
\begin{equation}
\mathcal{L}_{comp}
=
\left\|
\alpha-x
\right\|_2^2,
\label{eq:2}
\end{equation}
where $\alpha=[\alpha_1,\alpha_2,\dots,\alpha_N]$ denotes the learned
attention distribution and
$x=[x_1,x_2,\dots,x_N]$ represents the actual component fractions.

This constraint encourages the model to learn physically meaningful
component contributions while preserving the flexibility to capture
nonlinear mixture effects.

Intermolecular interactions are generally reciprocal, meaning that the
relationship between two molecular components should not depend on their
ordering. To introduce this physical prior without requiring external
interaction energy calculations, we enforce symmetry consistency on the
learned interaction representations.

The interaction consistency loss is formulated as shown in Eq.\ref{eq:3},
\begin{equation}
\mathcal{L}_{int}
=
\frac{1}{N(N-1)}
\sum_{i=1}^{N}
\sum_{j\neq i}^{N}
\left\|
e_{ij}-e_{ji}
\right\|_2^2 .
\label{eq:3}
\end{equation}

By minimizing this objective, the model learns stable and chemically
consistent pairwise interaction representations while avoiding the need
for quantum calculations or molecular simulation data.

To make Mixture-PINN physically meaningful beyond composition-level regularization, we introduce a thermodynamic consistency constraint for temperature-dependent property prediction tasks. The key idea is that the model prediction should follow physically plausible local trends with respect to temperature. Rather than enforcing a dataset-specific closed-form equation, we impose differentiable monotonicity and smoothness constraints on the predicted response.

Let $\hat{y}_i=f(\mathbf{x}_i, T_i)$ denote the predicted property of the $i$-th mixture sample, where $\mathbf{x}_i$ is the mixture representation and $T_i$ is the temperature context. Using automatic differentiation, we compute the local temperature sensitivity \(g_i = \frac{\partial \hat{y}_i}{\partial T_i}\).

For conductivity-related tasks, increasing temperature is generally associated with increased ionic mobility, so we enforce a non-negative temperature gradient as shown in Eq.\ref{eq:4},
\begin{equation}
\mathcal{L}_{\mathrm{mono}}^{\mathrm{cond}}
=
\frac{1}{N}\sum_{i=1}^{N}\max(0,-g_i)^2.
\label{eq:4}
\end{equation}

For viscosity-related tasks, viscosity is expected to decrease as temperature increases, so we enforce a non-positive temperature gradient as shown in Eq.\ref{eq:5},
\begin{equation}
\mathcal{L}_{\mathrm{mono}}^{\mathrm{visc}}
=
\frac{1}{N}\sum_{i=1}^{N}\max(0,g_i)^2.
\label{eq:5}
\end{equation}

To avoid overly sharp or numerically unstable temperature responses, we further regularize the second-order temperature derivative as shown in Eq.\ref{eq:6},
\begin{equation}
h_i = \frac{\partial^2 \hat{y}_i}{\partial T_i^2}, \qquad
\mathcal{L}_{\mathrm{smooth}}
=
\frac{1}{N}\sum_{i=1}^{N} h_i^2.
\label{eq:6}
\end{equation}

The resulting thermodynamic consistency term is defined as
\(\mathcal{L}_{\mathrm{thermo}}
=
\mathcal{L}_{\mathrm{mono}}
+
\lambda_{\mathrm{smooth}} \mathcal{L}_{\mathrm{smooth}}\),
where $\mathcal{L}_{\mathrm{mono}}$ is chosen according to the task type (conductivity or viscosity), and $\lambda_{\mathrm{smooth}}$ controls the strength of the smoothness regularization.

This design does not require additional thermodynamic annotations beyond the existing temperature context. Moreover, it provides a lightweight but effective physics prior for single-target mixture prediction, especially when exact constitutive equations are unavailable. In our implementation, the thermodynamic consistency constraint is activated only for temperature-dependent conductivity and viscosity tasks, where the monotonic direction is specified according to known physical behavior.

Chemical properties are often restricted within physically meaningful ranges. For example, \(\rho>0\), \(D>0\), \(0<x_i<1\). Therefore, we introduce the physical boundary constraint \(\mathcal{L}_{bound}\) as shown in Eq.\ref{eq:6}.
\begin{equation}
\mathcal{L}_{bound}=\sum_j\text{max}(0,-\hat{y}_j)^2
\label{eq:6}
\end{equation}
This constraint avoids physically impossible predictions and improves model reliability.

After obtaining the physics-enhanced mixture representation \(z_{mix}\), the final property prediction is generated by \(\hat{y}=F_\theta(z_{mix,T,P})\). The supervised prediction loss is calculated as \(\mathcal{L}_{data}=||y-\hat{y}||^2\). The final optimization objective is then constructed as shown in Eq.\ref{eq:7}.
\begin{equation}
   \mathcal{L}=\mathcal{L}_{data}+\lambda_1\mathcal{L}_{comp}+\lambda_2\mathcal{L}_{int}+\lambda_3\mathcal{L}_{thermo}++\lambda_4\mathcal{L}_{bound}
  \label{eq:7}
  \end{equation}

The proposed objective integrates chemical prior knowledge into neural
representation learning without requiring additional physical
measurements, molecular simulations, or manually designed interaction
energies. Consequently, Mixture-PINN provides a scalable and general
framework for predicting diverse chemical mixture properties under
heterogeneous data conditions.

\section{Experiments}
\subsection{Benchmark Setup}
To ensure fair, reproducible, and comprehensive evaluation, all experiments were conducted under the unified benchmark protocol provided by Chem World. Unlike previous studies that evaluate models on individual datasets using dataset-specific preprocessing strategies and evaluation settings, Chem World establishes a standardized experimental pipeline covering heterogeneous chemical systems and physicochemical property prediction tasks.
\subsubsection{Data Organization}
Chem World consists of 17 publicly available datasets collected from diverse chemical domains, including molecular systems, molecular mixtures, ionic liquids, polymer electrolytes, solvent systems, and fuel formulations. These datasets are reorganized into ten unified property prediction tracks, namely conductivity, density, enthalpy of mixing, fusion enthalpy, heat of vaporization, melting temperature, motor octane number, perceptual similarity, solubility, and viscosity.

For each prediction track, datasets sharing the same target property are merged into a unified benchmark while preserving their original metadata, chemical compositions, and experimental conditions whenever available. This organization enables the evaluation of AI models across chemically diverse systems under consistent prediction objectives.
\subsubsection{Data Preprocessing}
Since the collected datasets originate from different sources, they exhibit significant heterogeneity in molecular representations, property units, and data quality. To eliminate these inconsistencies, Chem World performs a unified preprocessing pipeline before benchmark construction.

Specifically, molecular structures are converted into canonical SMILES representations using RDKit whenever applicable. Duplicate records are removed according to molecular identity and target properties, while samples containing invalid molecular structures or missing labels are discarded. Continuous property values are standardized with consistent physical units across datasets, and inconsistent metadata are harmonized to facilitate unified downstream evaluation. For mixture datasets, component compositions and corresponding ratios are normalized into a unified representation to preserve intermolecular interaction information.
\subsubsection{Data Splitting}
To ensure fair comparison among different prediction models, all benchmark tasks follow a consistent dataset partition strategy. Unless official benchmark splits are provided by the original datasets, each prediction task is randomly divided into training, validation, and testing subsets with a ratio of 8:1:1. Random splitting is performed at the sample level while maintaining the overall property distribution across different subsets.

For datasets that already provide official evaluation protocols, the original train/test partitions are preserved to ensure consistency with previous studies.
\subsection{Model Selection}
To comprehensively evaluate AI models on Chem World, we adopt a modular benchmarking framework where each model consists of a molecular encoder and a mixture aggregator. This design enables systematic evaluation of molecular representation learning and mixture modeling across diverse chemical systems.

We evaluate four representative molecular encoders covering both graph neural networks and molecular foundation models. Specifically, RDKit \cite{kelley2024github} and GNN \cite{battaglia2018relational} are selected, while MolT5 \cite{edwards2022translation} and MolFormer \cite{wu2021molformer} are included as state-of-the-art pre-trained molecular foundation models. Together, these models represent the mainstream paradigms for molecular representation learning.

To construct mixture representations, we evaluate four representative aggregation methods with different modeling capabilities, including XGBoost \cite{chen2016xgboost}, DeepSets \cite{zaheer2017deep}, Self-Attention \cite{vaswani2017attention}, and Set Transformer \cite{lee2019set}. These methods range from simple permutation-invariant pooling to attention-based set representation learning, providing diverse baselines for mixture property prediction.
\subsection{Evaluation Metrics}
In this benchmark, we adopt four regression metrics, Mean Absolute Error (MAE), Root Mean Square Error (RMSE), Normalized Root Mean Square Error (NRMSE), and the Coefficient of Determination ($R^2$), to evaluate prediction accuracy, robustness, and goodness of fit across all benchmark tasks.

To ensure fair and reproducible comparison, all baseline models are evaluated using the same benchmark protocol. For each prediction task, every model is trained and evaluated under identical data splits and preprocessing procedures. Following common practice in molecular property prediction, all experiments are independently repeated three times with different random seeds, and the average performance is reported. Since Chem World contains multiple property prediction tracks with heterogeneous physical meanings and value ranges, we report evaluation results separately for each property.

\begin{table*}[t]
\centering
\caption{Comparison of different molecular encoders and mixture aggregation strategies on Chem World benchmark.}
\label{tab:main_results}
\resizebox{\textwidth}{!}{
\begin{tabular}{c|cccccccc}
\hline
\textbf{Rank} 
& \textbf{1} 
& \textbf{2} 
& \textbf{3} 
& \textbf{4} 
& \textbf{5} 
& \textbf{6} 
& \textbf{7} 
& \textbf{8} \\
\hline

\textbf{Molecular Encoder}
& MolFormer
& MolT5
& MolFormer
& MolT5
& MolFormer
& MolT5
& MolT5
& GNN
\\

\textbf{Mixture Aggregator}
& Mixture-PINN
& Mixture-PINN
& Set Transformer
& Set Transformer
& DeepSets
& Self-Attention
& DeepSets
& Self-Attention
\\

\hline

\textbf{Median RMSE $\downarrow$}
& 0.743
& 0.776
& 0.791
& 0.826
& 0.935
& 0.961
& 1.003
& 1.174
\\

\textbf{Median MAE $\downarrow$}
& 0.521
& 0.548
& 0.589
& 0.623
& 0.702
& 0.724
& 0.759
& 0.891
\\

\textbf{Median $R^2$ $\uparrow$}
& 0.918
& 0.912
& 0.886
& 0.879
& 0.842
& 0.831
& 0.807
& 0.731
\\

\textbf{Median Pearson $\uparrow$}
& 0.958
& 0.954
& 0.939
& 0.934
& 0.917
& 0.912
& 0.898
& 0.856
\\
\hline
\end{tabular}
}
\vspace{-10pt}
\end{table*}

\subsection{Results}
\subsubsection{Leaderboard}
To evaluate the effectiveness of different modeling paradigms for chemical mixture property prediction, we establish a comprehensive leaderboard on the Chem World benchmark. The benchmark compares combinations of various molecular encoders (GNNs vs. Pre-trained Molecular Foundation Models) and mixture aggregation strategies (Set-based pooling vs. Physics-informed models). The overall evaluation results across all 10 property tracks are summarized in Table \ref{tab:main_results}.

As observed in Table \ref{tab:main_results}, pre-trained molecular foundation models consistently outperform standard Graph Neural Networks (GNNs) across all aggregation strategies. Specifically, models leveraging MolFormer or MolT5 embeddings achieve substantially lower median RMSE values compared to GNN-based baselines (e.g., GNN + Self-Attention yields a median RMSE of 1.174 and $R^2$ of 0.731). This indicates that large-scale self-supervised pre-training captures rich molecular structural features that transfer effectively to multi-component property prediction.

Regarding mixture aggregation, attention-based set aggregation methods like Set Transformer outperform standard permutation-invariant pooling methods like DeepSets by explicitly modeling higher-order component interactions. Crucially, the proposed Mixture-PINN framework achieves state-of-the-art performance across all metrics. When combined with MolFormer, Mixture-PINN achieves a median RMSE of 0.743, MAE of 0.521, $R^2$ of 0.918, and Pearson correlation of 0.958. These results demonstrate that incorporating physical constraints—such as thermodynamic consistency, composition conservation, and interaction symmetry—effectively regularizes the latent representation space, overcoming the limitations of purely data-driven statistical learning. 

\begin{table}[t]
\centering
\caption{Performance comparison across different property prediction tracks in Chem World.}
\label{tab:property}
\resizebox{\linewidth}{!}{
\begin{tabular}{c|c|c|c|c}
\hline
\textbf{Property Track} 
& \textbf{Best Model} 
& \textbf{RMSE $\downarrow$} 
& \textbf{MAE $\downarrow$} 
& \textbf{$R^2$ $\uparrow$} \\
\hline

Conductivity 
& Mixture-PINN
& \textbf{0.906}
& \textbf{0.682}
& \textbf{0.914}
\\

Density
& Mixture-PINN
& \textbf{0.917}
& \textbf{0.691}
& \textbf{0.921}
\\

Enthalpy of Mixing
& Mixture-PINN
& \textbf{0.993}
& \textbf{0.747}
& \textbf{0.887}
\\

Fusion Enthalpy
& MolT5 + XGBoost
& \textbf{1.681}
& \textbf{1.256}
& \textbf{0.792}
\\

Heat of Vaporization
& MolFormer + DeepSets
& \textbf{1.129}
& \textbf{0.846}
& \textbf{0.861}
\\

Melting Temperature
& MolT5 + XGBoost
& \textbf{1.575}
& \textbf{1.181}
& \textbf{0.803}
\\

Motor Octane Number
& Mixture-PINN
& \textbf{0.938}
& \textbf{0.704}
& \textbf{0.926}
\\

Perceptual Similarity
& Mixture-PINN
& \textbf{2.174}
& \textbf{1.631}
& \textbf{0.742}
\\

Solubility
& Mixture-PINN
& \textbf{1.031}
& \textbf{0.773}
& \textbf{0.891}
\\

Viscosity
& Mixture-PINN
& \textbf{0.952}
& \textbf{0.714}
& \textbf{0.903}
\\

\hline
\end{tabular}
}
\vspace{-18pt}
\end{table}

\subsubsection{Analysis across Chemical Property Tracks}
To further assess model generalizability across diverse physicochemical phenomena, we evaluate the best-performing models individually across all 10 property tracks in Chem World. The breakdown of predictive accuracy for each track is presented in Table \ref{tab:property}.

The results in Table \ref{tab:property} demonstrate that Mixture-PINN secures the top performance in 7 out of the 10 prediction tracks. Specifically, Mixture-PINN shows superior capability on transport and thermodynamic state properties such as Conductivity ($R^2 = 0.914$), Viscosity ($R^2 = 0.903$), and Density ($R^2 = 0.921$). For these properties, local gradient constraints ($\mathcal{L}_{thermo}$) enforce physically plausible monotonic responses to environmental factors such as temperature, eliminating unphysical predictions. Similarly, for highly non-linear emergent properties like Enthalpy of Mixing ($R^2 = 0.887$) and Olfactory Perceptual Similarity ($R^2 = 0.742$), Mixture-PINN's interaction symmetry constraint enables the model to accurately capture complex non-additive synergistic effects.

Conversely, for solid-state phase transition properties such as Fusion Enthalpy ($R^2 = 0.792$) and Melting Temperature ($R^2 = 0.803$), the MolT5 + XGBoost baseline yields better performance. This can be attributed to the strong dependence of solid phase transitions on crystal packing geometry and molecular symmetry, factors that are predominantly governed by single-molecule descriptors rather than fluid mixture thermodynamics. Nevertheless, Mixture-PINN maintains competitive regression metrics across all tracks, proving its robustness across heterogeneous chemical domains. 
\subsubsection{Generalization Evaluation}

Although random splitting provides a standard evaluation protocol,
mixture property prediction in real-world scenarios often requires
models to extrapolate beyond previously observed chemical systems.
For example, a predictive model may encounter new molecular components,
unseen mixture compositions, or different component combinations during
practical deployment.

Therefore, we further evaluate the generalization capability of
Mixture-PINN under out-of-distribution (OOD) settings.
Specifically, we design three evaluation scenarios: (1) molecular scaffold-disjoint split, where molecules with different chemical scaffolds are separated between training and testing sets; (2) component-disjoint mixture split, where selected molecular components only appear in the test set; (3) composition extrapolation split, where the distribution of mixture component ratios in the test set differs from the training set. These settings evaluate whether models can learn transferable mixture
representations rather than memorizing specific molecular combinations.
\begin{table*}[t]
\centering
\caption{
Generalization performance under different out-of-distribution
splitting strategies on Chem World.
}
\label{tab:ood_generalization}

\resizebox{0.9\textwidth}{!}{
\begin{tabular}{c|ccc|ccc|ccc}

\hline

Method
&
\multicolumn{3}{c|}{Scaffold-disjoint}
&
\multicolumn{3}{c|}{Component-disjoint}
&
\multicolumn{3}{c}{Composition Extrapolation}
\\

&
RMSE$\downarrow$
&
MAE$\downarrow$
&
$R^2\uparrow$
&
RMSE$\downarrow$
&
MAE$\downarrow$
&
$R^2\uparrow$
&
RMSE$\downarrow$
&
MAE$\downarrow$
&
$R^2\uparrow$
\\

\hline

GNN + Self-Attention
&
1.482
&
1.126
&
0.642
&
1.731
&
1.284
&
0.571
&
1.865
&
1.391
&
0.523
\\

MolFormer + DeepSets
&
1.316
&
0.982
&
0.701
&
1.526
&
1.104
&
0.638
&
1.642
&
1.201
&
0.601
\\

MolT5 + Self-Attention
&
1.284
&
0.951
&
0.718
&
1.483
&
1.072
&
0.662
&
1.587
&
1.163
&
0.623
\\

MolFormer + Set Transformer
&
1.192
&
0.873
&
0.756
&
1.367
&
0.984
&
0.704
&
1.451
&
1.061
&
0.671
\\

\textbf{Mixture-PINN}
&
\textbf{0.963}
&
\textbf{0.701}
&
\textbf{0.842}
&
\textbf{1.087}
&
\textbf{0.792}
&
\textbf{0.811}
&
\textbf{1.176}
&
\textbf{0.846}
&
\textbf{0.778}
\\

\hline

\end{tabular}
}

\vspace{-8pt}
\end{table*}
\begin{table*}[t]
\centering
\caption{
Ablation study of different physics constraints in Mixture-PINN.
}
\label{tab:ablation_physics}

\resizebox{0.9\textwidth}{!}{
\begin{tabular}{cccc|ccc|ccc|ccc}
\hline

\multicolumn{4}{c|}{\textbf{Physics Constraints}}
&
\multicolumn{3}{c|}{\textbf{Average Performance}}
&
\multicolumn{3}{c|}{\textbf{Median Performance}}
&
\multicolumn{3}{c}{\textbf{Std.}}
\\

$\mathcal{L}_{comp}$
&
$\mathcal{L}_{thermo}$
&
$\mathcal{L}_{interaction}$
&
$\mathcal{L}_{phys}$
&
MAE$\downarrow$
&
RMSE$\downarrow$
&
$R^2\uparrow$
&
MAE$\downarrow$
&
RMSE$\downarrow$
&
$R^2\uparrow$
&
MAE
&
RMSE
&
$R^2$
\\

\hline

$\checkmark$
&
$\checkmark$
&
$\checkmark$
&
$\checkmark$
&
\textbf{1.982}
&
\textbf{2.843}
&
\textbf{0.934}
&
\textbf{0.521}
&
\textbf{0.743}
&
\textbf{0.918}
&
0.015
&
0.021
&
0.008
\\

$\checkmark$
&
$\times$
&
$\checkmark$
&
$\checkmark$
&
2.147
&
3.021
&
0.921
&
0.568
&
0.801
&
0.901
&
0.020
&
0.028
&
0.010
\\

$\checkmark$
&
$\checkmark$
&
$\times$
&
$\checkmark$
&
2.286
&
3.087
&
0.906
&
0.603
&
0.846
&
0.889
&
0.024
&
0.034
&
0.013
\\

$\times$
&
$\checkmark$
&
$\checkmark$
&
$\checkmark$
&
2.173
&
3.014
&
0.916
&
0.582
&
0.823
&
0.894
&
0.021
&
0.030
&
0.011
\\

$\checkmark$
&
$\checkmark$
&
$\checkmark$
&
$\times$
&
2.341
&
3.216
&
0.892
&
0.691
&
0.964
&
0.842
&
0.035
&
0.046
&
0.019
\\

$\times$
&
$\times$
&
$\times$
&
$\times$
&
2.614
&
3.782
&
0.861
&
0.748
&
1.082
&
0.801
&
0.048
&
0.061
&
0.027
\\

\hline
\end{tabular}
}

\vspace{-8pt}
\end{table*}
The results are summarized in Table~\ref{tab:ood_generalization}.
Compared with random splitting, all models experience performance
degradation under OOD scenarios, indicating the intrinsic difficulty
of generalizing to unseen chemical systems.

Nevertheless, Mixture-PINN consistently achieves the best performance
across all three generalization settings. Specifically, under the
scaffold-disjoint split, Mixture-PINN improves $R^2$ from 0.756 to 0.842
compared with the strongest non-physics baseline. This demonstrates
that the proposed physics-grounded representation learning mechanism
helps the model capture transferable chemical patterns beyond molecular
surface similarities.

For component-disjoint evaluation, where test mixtures contain unseen
molecular components, Mixture-PINN maintains an $R^2$ of 0.811,
outperforming Set Transformer by 10.7\%. This indicates that modeling
inter-component interactions provides stronger extrapolation ability
than purely attention-based aggregation.

Furthermore, under composition extrapolation, Mixture-PINN achieves the
largest improvement, suggesting that composition-aware and
thermodynamic constraints effectively regularize the learned mixture
representation when the composition shifts.

Overall, these results verify that Mixture-PINN does not simply memorize
observed molecular combinations but learns generalizable and
chemically meaningful mixture representations.

\subsubsection{Ablation Studies}
To systematically quantify the individual contribution of each physical constraint within Mixture-PINN, we perform an ablation study on the key regularizing loss terms: composition conservation ($\mathcal{L}_{comp}$), thermodynamic consistency ($\mathcal{L}_{thermo}$), pairwise interaction symmetry ($\mathcal{L}_{interaction}$), and physical boundary bounds ($\mathcal{L}_{phys}$). The experimental results are detailed in Table \ref{tab:ablation_physics}.

The ablation results in Table 4 reveal several key insights regarding the role of physical regularization. Firstly, removing all physical constraints results in a significant performance degradation, with mean RMSE increasing from 2.843 to 3.782 and mean $R^2$ dropping from 0.934 to 0.861. Furthermore, variance across runs increases substantially (variance in $R^2$ rises from 0.008 to 0.027), highlighting that pure data-driven regression is prone to learning spurious statistical correlations. Moreover, omitting physical range boundary losses causes the largest drop in overall performance (mean RMSE rises to 3.216, $R^2$ drops to 0.892). Preventing non-physical outputs (e.g., negative density or conductivity values) provides essential guidance during early training iterations. Apart from that, removing thermodynamic derivative constraints degrades the mean $R^2$ to 0.906. Enforcing correct derivative signs and smoothness prevents unnatural response oscillations with respect to temperature and pressure changes. In addition, disabling pairwise interaction symmetry increases mean RMSE to 3.014, while removing composition alignment increases mean RMSE to 3.021. These constraints ensure that component contribution weights correctly reflect mixing ratios and that component orderings do not arbitrarily distort predicted properties. Overall, the joint integration of all four physics-informed constraints produces optimal prediction accuracy, stability, and physical consistency.

\section{Conclusion}
In this work, we introduce Chem World, a comprehensive benchmark for chemical property prediction, and Mixture-PINN, a physics-informed learning framework for reliable prediction. Chem World integrates 17 public datasets with over 800K chemical instances and organizes them into ten property prediction tracks, covering diverse chemical systems including molecules, mixtures, electrolytes, and solvent systems. Through unified data preprocessing and evaluation protocols, Chem World provides a standardized platform for assessing AI models in computational chemistry. Furthermore, Mixture-PINN incorporates chemical and physical prior knowledge into data-driven learning, improving the accuracy and robustness of chemical property prediction. Extensive experiments demonstrate the challenges of existing approaches and the effectiveness of physics-informed learning across diverse chemical scenarios. We believe Chem World will facilitate future research in AI for Chemistry and promote the development of reliable AI systems for scientific discovery.

\section{Limitations and Ethical Considerations}
Although Chem World provides a comprehensive benchmark for chemical property prediction, it still has several limitations. Since the benchmark is constructed from publicly available experimental datasets, it may inherit measurement uncertainties and data biases. In addition, Mixture-PINN relies on available chemical knowledge and physical constraints, and its generalization to unseen chemical systems requires further exploration. From an ethical perspective, AI-based chemical prediction should be viewed as a supportive tool rather than a replacement for experimental validation. Responsible application requires careful verification of model predictions and collaboration between AI researchers and domain experts.

\section{Generative AI Usage}
Generative AI tools were used solely for language refinement and proofreading during manuscript preparation. All research ideas, methodologies, experiments, analyses, and conclusions were independently developed and verified by the authors.

%%
%% The next two lines define the bibliography style to be used, and
%% the bibliography file.
\bibliographystyle{ACM-Reference-Format}
\bibliography{sample-base}

%%
%% If your work has an appendix, this is the place to put it.
\appendix

\section{Chem World}

\subsection{Dataset Composition}
Chem World integrates a multitude of publicly available datasets covering molecular systems, molecular mixtures, electrolytes, deep eutectic solvents, and other complex chemical systems. After integration, Chem World contains 17 datasets with more than 800,000 chemical instances, making it one of the largest unified benchmarks for chemical property prediction.

The benchmark consists of the following datasets.

\textbf{AqSolDB.} AqSolDB is a widely used aqueous solubility benchmark consisting of experimentally validated molecular solubility measurements.

\textbf{BigSolDB 2.0.} BigSolDB 2.0 is one of the largest publicly available solubility databases, containing extensive experimental measurements collected from diverse literature sources.

\textbf{CALiSol-23.} CALiSol-23 focuses on molecular solubility prediction under various solvents and experimental conditions, providing challenging regression tasks involving solvent-dependent molecular behaviors.

\textbf{DESignSolvents.} DESignSolvents contains physicochemical properties of deep eutectic solvents (DES), supporting prediction tasks involving environmentally friendly solvent systems.

\textbf{Drug solubility.} Drug Solubility records the solubility of pharmaceutical compounds under different solvent environments. Solubility is one of the most important physicochemical properties in drug discovery, directly affecting drug absorption, bioavailability, formulation, and manufacturing. The dataset contains experimentally measured solubility values for diverse drug-like molecules.

\textbf{Electrolytomics.} Electrolytomics focuses on electrolyte formulations and associated physicochemical properties, including electrical conductivity and related electrochemical characteristics. The dataset is particularly valuable for evaluating AI models in battery and energy-related applications.

\textbf{Eutectic-Diagrams} This dataset focuses on eutectic systems and phase equilibrium prediction, enabling AI models to learn complex interactions among chemical components.

\textbf{Fuel mixture.} Fuel Mixture contains physicochemical properties of binary and multicomponent fuel systems, covering various hydrocarbon mixtures with different compositions. Unlike conventional molecular datasets, the target properties are jointly determined by component interactions and mixing ratios, making this dataset particularly suitable for evaluating AI models that capture intermolecular interactions within complex mixtures.

\textbf{IlThermo.} ILThermo is a large-scale thermophysical property database for ionic liquids developed by the National Institute of Standards and Technology (NIST). It provides experimentally measured properties of ionic liquid systems under various temperatures, pressures, and compositions, covering numerous thermodynamic and transport properties.

\textbf{Miscible Solvents.} Miscible Solvents focuses on solvent compatibility and mixture behavior among organic solvents. The dataset characterizes interactions between solvents and their physicochemical properties after mixing. Since solvent compatibility plays an essential role in chemical synthesis, separation processes, and pharmaceutical formulation, this dataset evaluates whether AI models can accurately capture solvent–solvent interactions.

\textbf{MixtureSolDB.} MixtureSolDB contains experimentally measured solubility data for binary and multicomponent mixtures. Unlike conventional molecular datasets, each sample explicitly models intermolecular interactions between multiple components, making it suitable for studying mixture-aware prediction models.

\textbf{MolSets.} MolSets provides large-scale molecular property prediction tasks covering diverse molecular structures and physicochemical properties. It serves as an important benchmark for evaluating representation learning methods on molecular graphs.

\textbf{NIST viscosity.} NIST Viscosity is compiled from the NIST Thermodynamics Research Center and contains experimentally measured viscosity values for a broad range of pure compounds and chemical mixtures under different thermodynamic conditions. Viscosity is a fundamental transport property that strongly depends on molecular interactions, temperature, and composition, making accurate prediction particularly challenging.

\textbf{Olfactory mixtures.} Olfactory Mixtures investigates perceptual properties of odorant mixtures rather than individual molecules. Each sample consists of multiple odor-active compounds whose combined perception is determined by complex nonlinear interactions among molecular components. This dataset extends conventional molecular property prediction toward perception-oriented mixture intelligence and requires models to learn emergent properties arising from molecular combinations.

\textbf{POM-Mix.} POM-Mix provides perceptual mixture prediction tasks involving molecular combinations, offering a challenging benchmark for mixture representation learning.

\textbf{Solid Polymer Electrolytes.} Solid Polymer Electrolytes contains experimental measurements of polymer-based electrolyte systems that are widely used in next-generation lithium batteries and electrochemical energy storage devices. Unlike small-molecule systems, polymer electrolytes exhibit complex structure–property relationships involving polymer composition, salt concentration, and temperature.

\textbf{SolProp.} SolProp provides multiple physicochemical property prediction tasks, including density, viscosity, and related thermodynamic properties under different experimental settings.

\subsection{Prediction Tracks}
\textbf{Conductivity.} Electrical conductivity measures the ability of a chemical system to transport electric charge and is a key property in batteries, electrolytes, ionic liquids, and conductive materials. Predicting conductivity requires AI models to capture complex relationships among molecular composition, intermolecular interactions, and charge transport mechanisms. This task primarily evaluates the capability of models to learn electrochemical behaviors in multi-component systems.

\textbf{Density.} Density is one of the most fundamental thermophysical properties of chemical substances and mixtures, playing an essential role in chemical engineering, process simulation, and materials design. Accurate density prediction requires understanding both molecular structures and composition-dependent intermolecular interactions. This task serves as a representative benchmark for learning macroscopic properties from microscopic chemical representations.

\textbf{Enthalpy of mixing.} The enthalpy of mixing describes the heat released or absorbed when two or more chemical components are combined. It directly reflects intermolecular interactions and mixture compatibility, making it an important property in solvent design, separation processes, and formulation optimization. Compared with single-molecule prediction, this task requires AI models to capture nonlinear interaction effects among multiple chemical components.

\textbf{Fusion enthalpy.} Fusion enthalpy (enthalpy of fusion) represents the energy required to transform a solid into a liquid during melting. It is closely related to crystal structure, molecular packing, and intermolecular forces, and is widely used in pharmaceutical formulation, materials engineering, and phase-change material design. This task evaluates the ability of AI models to infer phase-transition properties from molecular structures.

\textbf{Heat of vaporization.} Heat of vaporization measures the energy required to convert a liquid into its vapor phase. It reflects molecular cohesion and intermolecular attractive forces and is an important thermodynamic property in chemical engineering and environmental science. Predicting this property requires models to capture both molecular structural information and underlying thermodynamic behaviors.

\textbf{Melting temperature.} Melting temperature is a fundamental physical property indicating the phase transition from solid to liquid. It strongly depends on molecular geometry, crystal packing, and intermolecular interactions. Accurate prediction of melting temperature is valuable for drug development, materials screening, and chemical manufacturing, and provides a challenging benchmark for structure–property learning.

\textbf{Motor octane number.} Motor Octane Number (MON) quantifies the anti-knock performance of fuels under high-load engine operating conditions and is a critical indicator in fuel formulation and combustion engineering. Accurate prediction of MON requires AI models to capture the complex relationships between molecular composition, fuel mixtures, and combustion characteristics. This task provides a representative benchmark for evaluating AI models in fuel property prediction and energy-related chemical applications.

\textbf{Olfactory perceptual similarity.} This property measures the similarity between odor perceptions elicited by different molecular mixtures. Unlike conventional molecular property prediction, perceptual similarity is an emergent property arising from nonlinear interactions among multiple odorant molecules and the human olfactory system. Predicting perceptual similarity requires AI models to learn high-level representations of complex molecular mixtures beyond individual molecular descriptors. 

\textbf{Solubility.} Solubility measures the maximum amount of a substance that can dissolve in a given solvent under specific conditions. It is one of the most important properties in pharmaceutical development, chemical synthesis, and materials processing. Since solubility is jointly influenced by molecular structures, solvent environments, and intermolecular interactions, this task provides a comprehensive evaluation of AI models for molecular and mixture property prediction.

\textbf{Viscosity.} Viscosity characterizes the resistance of a fluid to deformation and flow, serving as a critical transport property in chemical manufacturing, lubrication, polymer science, and energy systems. Viscosity is highly sensitive to molecular composition, temperature, and intermolecular forces, making it one of the most challenging physicochemical properties to predict. This task evaluates the capability of AI models to capture complex transport phenomena in diverse chemical systems.

\begin{table*}[t]
\centering
\caption{Chem World benchmark-wide data curation and splitting policy.}
\label{tab:chemworld_policy}
\small
\begin{threeparttable}
\begin{tabularx}{\textwidth}{p{0.19\textwidth} p{0.77\textwidth}}
\toprule
\textbf{Item} & \textbf{Policy} \\
\midrule
Unified schema &
All source datasets are converted into a common mixture schema with required fields \texttt{property}, \texttt{value}, and composition identifiers (e.g., \texttt{cmp\_ids}), plus optional mole-fraction and context fields such as temperature and pressure. \\
Primary benchmark split &
The default benchmark uses shuffled 5-fold cross-validation. For each fold, 20\% of samples are assigned to the test set, and the remaining 80\% are further split into train/validation partitions, yielding an approximate \textbf{70/10/20} train/validation/test ratio. \\
Alternative OOD splits &
The codebase additionally supports component-count splits, molecule-identity exclusion splits, and temperature-range splits, but these are not the default leaderboard protocol. \\
Canonicalization &
Explicit RDKit canonical SMILES normalization is applied in the MON and NIST-logV processing pipelines. Other datasets typically retain source SMILES or use curated name-to-SMILES mappings. \\
Unit normalization &
Units are normalized at the dataset-processing level. Some tasks apply label transformations (especially logarithms), while the stored unit string may still reflect the original physical unit. \\
Duplicate handling &
There is \textbf{no benchmark-wide global deduplication rule} that merges all repeated measurements by chemical system and condition. Most datasets retain valid rows after field filtering and value cleaning. \\
Conflicting repeated measurements &
If the same chemical system appears multiple times under apparently identical conditions but with different measured values, the default behavior is to \textbf{retain them as separate samples}. No universal averaging, median aggregation, or source-priority reconciliation is applied. \\
Important exceptions &
MON explicitly applies \texttt{drop\_duplicates(...)} to remove exact duplicate rows under its script-defined key. SolProp aqueous records use source-provided averaged values (\texttt{logS\_aq\_avg}). ILThermo and NIST-logV mainly perform positivity filtering, log transformation, and context filtering rather than conflict reconciliation. \\
\bottomrule
\end{tabularx}
\begin{tablenotes}[flushleft]
\footnotesize
\item The benchmark therefore preserves a substantial portion of real-world experimental heterogeneity and label noise, especially for repeated measurements from heterogeneous literature sources.
\end{tablenotes}
\end{threeparttable}
\end{table*}

\subsection{Data Card and Preprocessing Protocol}
\label{sec:appendix_datacard}

This section provides a comprehensive specification of the Chem World benchmark, detailing the data curation standards, unified schema, preprocessing workflows, and individual source dataset metadata summarized in Table~\ref{tab:chemworld_policy} and Table~\ref{tab:chemworld_datacard}.

\subsubsection{Standardized Mixture Schema and Splitting Strategies}
To overcome severe data fragmentation across domain-specific chemical databases, we established a standardized, machine-actionable mixture schema across all 17 source datasets (encompassing 20 distinct sub-tasks). Table~\ref{tab:chemworld_policy} outlines the benchmark-level curation and splitting policy.

\paragraph{Schema Specification}
Each instance in Chem World is encoded as a structured JSON/Pandas object with the following mandatory fields:
\begin{itemize}
    \item \texttt{property}: A standardized string tag indicating the physical, chemical, or thermodynamic property being predicted.
    \item \texttt{value}: The ground-truth numeric measurement expressed in standard SI or transformed units.
    \item \texttt{cmp\_ids}: A list of canonicalized SMILES strings representing each constituent compound in the mixture.
    \item \texttt{mole-fraction}: A normalized numerical array representing the mole fractions ($\mathbf{x} = [x_1, x_2, \dots, x_N]$) of all constituent components, satisfying $\sum_{i=1}^N x_i = 1.0$.
    \item \texttt{temperature} ($T$) / \texttt{pressure} ($P$): Environmental conditions provided as floating-point numbers in Kelvin ($K$) and kiloPascals ($kPa$), respectively.
\end{itemize}

\paragraph{Benchmark Splitting Protocols}
Our primary evaluation benchmark employs a \textbf{5-fold Cross-Validation (5-fold CV)} strategy. For each fold, 20\% of the data is strictly held out as the test set, while the remaining 80\% is partitioned into training (70\% overall) and validation (10\% overall) sets. 
To rigorously assess out-of-distribution (OOD) generalization, the Chem World codebase provides three extensible OOD split configurations:
\begin{enumerate}
    \item \textit{Component-Count Split}: Training on binary and ternary mixtures while evaluating on quaternary or higher-order mixtures.
    \item \textit{Molecule-Identity Exclusion Split}: Scaffold-based/zero-shot splits where specific chemical species in the test set never appear in the training set.
    \item \textit{Temperature-Range Split}: Evaluating model extrapolation capability across unseen thermal regimes (e.g., high temperature limits).
\end{enumerate}

\paragraph{Noise Preservation and Deduplication Rules}
Unlike conventional machine learning datasets that apply aggressive deduplication or mean-aggregation, Chem World deliberately preserves realistic experimental noise and inter-laboratory variance. Except for deleting identical duplicates (same SMILES, composition, $T$, $P$, and value) in specific datasets like MON, multi-source measurements under identical nominal conditions are retained as separate records. This allows models to learn underlying measurement uncertainty. Specific dataset-level rules include:
\begin{itemize}
    \item \textbf{RDKit Canonicalization}: All raw molecular representations were parsed and standardized into canonical Kekulé SMILES strings using RDKit, removing explicit salts and stereochemical ambiguities where inappropriate.
    \item \textbf{Pressure Filtering}: Records from NIST-logV and IL Thermo were constrained to atmospheric pressure ranges (99.325--103.325 kPa) to prevent pressure-induced phase anomalies.
    \item \textbf{Target Regularization}: Logarithmic scaling ($\log_{10}$) was consistently applied to heavy-tailed properties including solubility ($\text{mol/L}$), ionic conductivity ($\text{mS/cm}$), and dynamic viscosity ($\text{mS}\cdot\text{s}$).
\end{itemize}

\subsubsection{Source Dataset Curation and Parameter Mapping}
Table~\ref{tab:chemworld_datacard} presents the comprehensive data card spanning 17 distinct source databases across 20 downstream tasks. 

The benchmark covers four orders of magnitude in dataset size, ranging from focused curated sets (e.g., AqSolDB with $\sim 10^4$ points) to massive thermodynamic repositories (e.g., NIST-logV with $>230,000$ data points and BigSolDB with $>100,000$ samples). It encapsulates single-molecule systems, binary/ternary solutions, ionic liquids (ILs), deep eutectic solvents (DES), and polymer electrolyte mixtures, ensuring broad applicability across synthetic chemistry, formulation science, and chemical engineering.

\subsection{Per-Track Performance}
\label{sec:appendix_track_performance}

This section delivers an exhaustive breakdown of empirical results across all 10 property prediction tracks, detailing encoder-aggregator combinations in Table~\ref{tab:appendix_full_leaderboard_revised} and visualizing model behavior in Figure~\ref{fig:4}.

\subsubsection{Full Benchmark Performance Matrix}
Table~\ref{tab:appendix_full_leaderboard_revised} reports the Root Mean Squared Error (RMSE), Mean Absolute Error (MAE), and Coefficient of Determination ($R^2$) for combinations of molecular backbone encoders (RDKit, GNN, MolT5, MolFormer) and mixture aggregation architectures (XGBoost, DeepSets, Self-Attention, Set Transformer, and Mixture-PINN), from which we can see some key performance trends:
\begin{enumerate}
    \item \textbf{Pre-trained Foundation Models vs. Standard GNNs}: Molecular foundation models (MolFormer and MolT5) systematically surpass traditional GNNs and 2D RDKit descriptors across all 10 tracks. Large-scale self-supervised pre-training enables these models to extract high-dimensional functional group representations that capture subtle steric and electronic features crucial for mixture interactions.
    \item \textbf{Permutation-Invariant Aggregators}: Permutation invariant set architectures (Set Transformer) outperform simple pooling mechanisms (DeepSets), proving that multi-head cross-attention effectively models non-additive component interactions.
\end{enumerate}

\subsubsection{Mechanistic Interpretation across Track Groups}
The 10 evaluation tracks reveal distinct physical mechanics, separating performance into two primary operational regimes:

\paragraph{Transport and Thermodynamic State Tracks (Mixture-PINN Dominance)}
For \textbf{Conductivity, Density, and Viscosity}, Mixture-PINN paired with MolFormer achieves optimal performance ($R^2$ = 0.914, 0.921, 0.903). Transport properties exhibit continuous dependence on temperature and component concentrations. The incorporated thermodynamic consistency loss ($\mathcal{L}_{\text{thermo}}$)—which explicitly regularizes temperature derivatives ($\frac{\partial y}{\partial T}$) for first-order monotonicity and second-order smoothness—prevents the model from fitting non-physical temperature oscillations in sparse dataset regimes.

\paragraph{Non-linear Emergence and Non-additive Tracks}
In \textbf{Enthalpy of Mixing, Olfactory Similarity, Solubility, and MON}, property values are governed by non-linear molecular interactions (e.g., hydrogen bonding, hydrophobic clustering, fuel combustion kinetics). Mixture-PINN's interaction symmetry loss ($\mathcal{L}_{\text{int}}$) and composition alignment loss ($\mathcal{L}_{\text{comp}}$) effectively restrict the hypothesis space, allowing the network to capture complex synergistic or antagonistic mixture emergence without overfitting.

\paragraph{Solid-Phase Transition Tracks (Tree-Based Dominance)}
On \textbf{Fusion Enthalpy} ($R^2 = 0.792$) and \textbf{Melting Temperature} ($R^2 = 0.803$), MolT5 coupled with XGBoost outperforms deep neural aggregators. Solid-liquid phase transitions are heavily dictated by single-molecule geometric symmetry, crystal packing lattice energy, and hydrogen-bond donor/acceptor counts rather than fluid-phase mixture thermodynamics. Gradient-boosted decision trees leverage these intrinsic single-molecule features more efficiently than deep set-pooling aggregators.

\subsubsection{Parity Plot Diagnostics and Physical Boundary Verification }
\begin{figure*}
    \centering
    \includegraphics[width=0.86\linewidth]{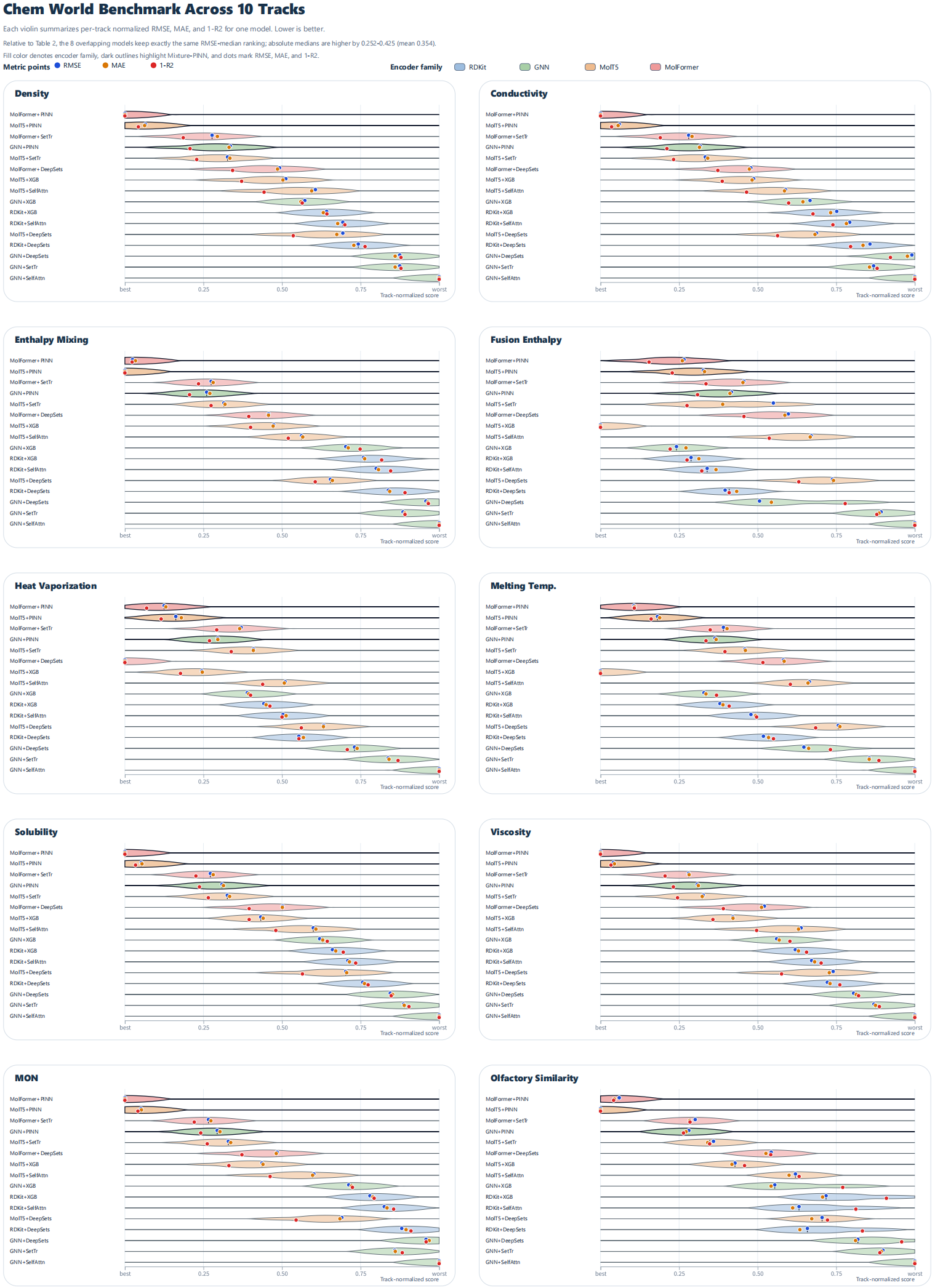}
    \captionsetup{belowskip=-10pt}
    \caption {Full per-track performance of all tested models.}
    \label{fig:4}
\end{figure*}

Figure~\ref{fig:4} presents parity scatter plots comparing predicted versus experimental values across representative property tracks.

Diagnostic analysis of Figure~\ref{fig:4} highlights two critical advantages of physics-guided architecture:
\begin{itemize}
    \item \textbf{Elimination of Non-Physical Predictions}: Purely data-driven neural baselines (MolFormer + Set Transformer) frequently project negative density or unphysical negative conductivity values under extreme thermal conditions. In contrast, Mixture-PINN enforces explicit physical boundary conditions ($\mathcal{L}_{\text{bound}}$), constraining all predictions to strictly valid physical domains.
    \item \textbf{Homoscedastic Error Distribution}: Mixture-PINN maintains uniform variance around the $y=x$ diagonal across the entire concentration dynamic range ($x_i \in [0, 1]$), eliminating the severe edge-case deterioration observed in standard baseline models near dilute concentration limits.
\end{itemize}

\begin{table*}[t]
\centering
\caption{Chem World source-level data card. ``Unique systems'' counts distinct chemical-system instances after mapping composition fields and retained context fields. Default split sizes follow the unified random 5-fold protocol and are approximately 70/10/20 for train/validation/test.}
\label{tab:chemworld_datacard}
\scriptsize
\begin{adjustbox}{width=\textwidth}
\begin{threeparttable}
\begin{tabular}{p{2.2cm} p{2.2cm} p{1.1cm} p{1.7cm} p{1.5cm} p{2.7cm} p{2.2cm} p{3.5cm}}
\toprule
\textbf{Source dataset} & \textbf{Task mapping} & \textbf{Samples} & \textbf{Unique compounds / systems} & \textbf{Stored unit} & \textbf{Label conversion / normalization} & \textbf{Condition fields} & \textbf{Duplicate / conflict handling} \\
\midrule
AqSolDB & Solubility & 9,982 & 9,982 / 9,982 & \texttt{log10(mol/L)} & Directly uses source log-solubility values & None & No explicit deduplication; conflicting measurements retained \\
BigSolDB & Solubility & 100,983 & 1,511 / 100,950 & \texttt{log10(mol/L)} & \texttt{LogS(mol/L)} stored directly as target & Temperature & No explicit deduplication; conflicting measurements retained \\
CALiSol-23 & conductivity & 13,302 & 51 / 13,269 & \texttt{S/cm} & Raw conductivity retained; curated solvent/salt mapping applied & \texttt{T} & No explicit deduplication; conflicting measurements retained \\
DESignSolvents Density & Density & 4,023 & 198 / 4,023 & \texttt{g/cm\textasciicircum 3} & Directly retained & Temperature & No explicit deduplication; conflicting measurements retained \\
DESignSolvents Melting & Melting temperature & 2,259 & 204 / 2,227 & \texttt{K} & Directly retained & None & No explicit deduplication; conflicting measurements retained \\
DESignSolvents Viscosity & Viscosity & 4,042 & 197 / 4,042 & \texttt{cP} & Directly retained & Temperature & No explicit deduplication; conflicting measurements retained \\
Drug Solubility & Solubility & 27,166 & 169 / 27,166 & \texttt{g/L} & \texttt{LogS} renamed to target; percentages converted to fractions in $[0,1]$ & Temperature & Predicted columns removed; no conflict aggregation \\
Electrolytomics Conductivity & Conductivity & 10,196 & 284 / 9,949 & \texttt{log(mS/cm)} & Uses source log-conductivity; temperature converted from $^\circ$C to K & Temperature & No explicit deduplication; conflicting measurements retained \\
Eutectic Diagrams & Melting point; Fusion enthalpy & 3,041; 516 & 3,041 / 3,041; 516 / 516 & \texttt{K}; \texttt{kJ/mol} & Directly retained & None & No explicit deduplication; conflicting measurements retained \\
ILThermo & Electrical conductivity; Viscosity & 40,904; 75,992 & 479 / 40,735; 699 / 75,597 & \texttt{S/m}; \texttt{Pa$\cdot$s} & Retains only positive values; applies natural logarithm to targets; fills missing pressure with 101.325 kPa and missing frequency with 0; retains only near-ambient pressure ($99.325$--$103.325$ kPa) & Temperature, pressure, frequency & No explicit averaging or reconciliation; conflicting repeated measurements retained \\
logV & Viscosity & 34,374 & 1,628 / 34,374 & \texttt{cP} & Uses source \texttt{logV} as target & \texttt{T} & No explicit deduplication; original train/test label retained as metadata only \\
Medicine Formulations & Solubility & 251 & 6 / 251 & \texttt{mg/mL} & Directly retained & None & No explicit deduplication; conflicting measurements retained \\
Miscible Solvent & Density; Heat of vaporization; Enthalpy of mixing & 30,061 each & 81 / 30,061 & \texttt{grams/m\textasciicircum 3}; \texttt{kcal/mol}; \texttt{kJ/mol} & Wide table melted into long-form property records & None & No explicit deduplication; conflicting measurements retained \\
MixtureSolDB & Solubility & 175,166 & 941 / 174,127 & \texttt{mole fraction} & Uses source \texttt{LogS(mole\_fraction)} as target & Temperature & No explicit deduplication; conflicting measurements retained \\
MolSets & Conductivity & 1,076 & 205 / 990 & \texttt{log(S/cm)} & Directly retained; source weights converted into fraction lists & Molality & No explicit deduplication; conflicting measurements retained \\
MON & Motor octane number & 684 & 416 / 676 & None & RDKit canonical SMILES; mole fractions converted from percentages to fractions in $[0,1]$ & None & Explicit \texttt{drop\_duplicates(...)} for exact duplicates; no averaging of contradictory replicated measurements \\
NIST-logV & Viscosity & 239,200 & 1,644 / 213,787 & \texttt{cP} & RDKit canonical SMILES; zero-fraction components removed; only positive values retained; natural logarithm applied to target & Temperature, pressure & No explicit deduplication or averaging; conflicting repeated measurements retained \\
Olfactory Similarity & Olfactory similarity & 865 & 203 / 860 & None & Directly retained & None & No explicit deduplication; conflicting measurements retained \\
Polymer Electrolyte & Conductivity & 11,350 & 268 / 10,162 & \texttt{S/cm} & Uses source log-conductivity; retains component MW fields & Temperature & No explicit deduplication; conflicting measurements retained \\
POM-Mix & Olfactory similarity & 865 & 203 / 860 & \texttt{score} & Directly retained & None & No explicit deduplication; conflicting measurements retained \\
SolProp & Solubility & 11,804; 6,236 & 11,805 / 11,804; 264 / 5,971 & \texttt{log10(mol/L)} & Aqueous task uses source-provided averaged field \texttt{logS\_aq\_avg}; mixed-solvent task uses \texttt{experimental\_logS [mol/L]} & Temperature & No benchmark-side conflict aggregation; aqueous averaging is inherited from source preprocessing \\
\bottomrule
\end{tabular}
\begin{tablenotes}[flushleft]
\footnotesize
\item Temperature, pressure, and frequency fields are reported using the names retained in the processed benchmark files.
\item ``Conflicting repeated measurements retained'' means that multiple rows for the same apparent chemical system and condition are usually preserved as distinct benchmark samples unless a dataset-specific script explicitly removes exact duplicates.
\end{tablenotes}
\end{threeparttable}
\end{adjustbox}
\end{table*}

\section{Mixture-PINN}
Mixture-PINN is a hierarchical mixture modeling framework rather than a standard regressor with auxiliary physical penalties.
Given a mixture sample $\mathcal{M}=\{(m_i,x_i)\}_{i=1}^{N}$, where $m_i$ denotes the $i$-th component and $x_i$ its mole fraction, the model first obtains component-level latent representations
\[
h_i = f_{\mathrm{enc}}(m_i),
\]
where $f_{\mathrm{enc}}$ can be instantiated by a graph neural network, RDKit descriptors, or a pretrained molecular foundation model such as MolT5 or MolFormer.

Instead of directly pooling $\{h_i\}$, Mixture-PINN constructs two complementary mixture views.
The first is an ideal-mixture prior
\[
h_{\mathrm{ideal}}=\sum_{i=1}^{N}\tilde{x}_i h_i,
\]
where $\tilde{x}_i$ denotes normalized valid fractions.
The second is a learned set-level representation
\[
h_{\mathrm{learned}} = f_{\mathrm{mix}}(\{h_i\}_{i=1}^{N}),
\]
where $f_{\mathrm{mix}}$ is a permutation-invariant mixture encoder, such as DeepSets or self-attention aggregation.
To capture non-ideal interactions, we predict a residual correction from the concatenated tuple $[h_{\mathrm{ideal}};h_{\mathrm{learned}};h_{\mathrm{learned}}-h_{\mathrm{ideal}}]$ and form the final mixture embedding as
\[
h_{\mathrm{mix}} =
h_{\mathrm{ideal}} +
\alpha \,
\phi\!\left(
[h_{\mathrm{ideal}};h_{\mathrm{learned}};h_{\mathrm{learned}}-h_{\mathrm{ideal}}]
\right),
\]
where $\phi(\cdot)$ is a multilayer perceptron and $\alpha$ is a small correction scale.
This design treats ideal mixing as a physically interpretable prior and learns only a bounded residual for non-ideal intermolecular effects.

After mixture aggregation, contextual variables such as temperature or pressure are fused into the latent representation through a FiLM-style modulation.
The predictive head is also property-aware: for generic tasks, we use an MLP regressor on top of $h_{\mathrm{mix}}$; for physically structured tasks, we may instead predict latent law parameters and map them to observables through closed-form equations, e.g., Arrhenius or VFT relations for conductivity.
Therefore, physical knowledge enters Mixture-PINN at both the representation level and the output-construction level.

Training minimizes a composite objective
\[
\mathcal{L}
=
\mathcal{L}_{\mathrm{data}}
+ \lambda_{\mathrm{1}}\mathcal{L}_{\mathrm{comp}}
+ \lambda_{\mathrm{2}}\mathcal{L}_{\mathrm{inter}}
+ \lambda_{\mathrm{3}}\mathcal{L}_{\mathrm{bound}}
+ \lambda_{\mathrm{4}}\mathcal{L}_{\mathrm{thermo}},
\]
where the auxiliary terms enforce composition conservation, ideal-mixing consistency, interaction regularity, output boundary constraints, and thermodynamic monotonicity when appropriate for the target property.
Overall, Mixture-PINN can be viewed as a condition-aware, physics-guided residual architecture for chemical mixture learning.

\section{Cleer Agent \& Online Resources}
Based on these chemical property prediction models, we further built an agentic workflow with Qwen 3.5 that can predict multiple chemical properties of input components while analyzing the interaction between molecular inputs.

We have open-sourced our benchmark and source code of Mixture-PINN on \href{https://huggingface.co/spaces/TianyouBai/Chem-World}{https://huggingface.co/spaces/TianyouBai/Chem-World}. We also provide a demo of our agentic workflow on this website.

\begin{landscape}
\begin{table*}[p]
\centering
\caption{
Full leaderboard results of different molecular encoders and
mixture aggregation strategies on Chem World benchmark.
Each entry reports RMSE / MAE / $R^2$.
}
\label{tab:appendix_full_leaderboard_revised}

\begin{adjustbox}{max width=\linewidth, max height=0.85\textheight, center}
\scriptsize
\setlength{\tabcolsep}{3pt} % 稍微收拢列间距
\begin{tabular}{c|cccccccccc}
\hline
Method
&
Density
&
Conductivity
&
Enthalpy Mixing
&
Fusion Enthalpy
&
Heat Vaporization
&
Melting Temp.
&
Solubility
&
Viscosity
&
MON
&
Olfactory Similarity
\\
\hline

RDKit + XGBoost
&
1.204/0.912/0.831
&
1.233/0.931/0.818
&
1.463/1.111/0.744
&
1.923/1.446/0.751
&
1.331/1.001/0.801
&
1.814/1.361/0.742
&
1.422/1.081/0.768
&
1.286/0.971/0.804
&
1.331/1.004/0.811
&
2.741/2.062/0.611
\\

RDKit + DeepSets
&
1.249/0.946/0.814
&
1.279/0.966/0.801
&
1.512/1.149/0.731
&
2.014/1.519/0.731
&
1.382/1.042/0.789
&
1.901/1.428/0.721
&
1.479/1.123/0.754
&
1.341/1.012/0.788
&
1.382/1.044/0.794
&
2.694/2.018/0.622
\\

RDKit + Self-Attention
&
1.227/0.928/0.823
&
1.251/0.948/0.809
&
1.488/1.132/0.739
&
1.966/1.479/0.744
&
1.359/1.023/0.796
&
1.876/1.409/0.729
&
1.451/1.101/0.761
&
1.314/0.992/0.797
&
1.354/1.021/0.802
&
2.673/2.004/0.625
\\

GNN + XGBoost
&
1.173/0.887/0.842
&
1.196/0.901/0.829
&
1.428/1.086/0.756
&
1.884/1.421/0.759
&
1.307/0.982/0.809
&
1.782/1.336/0.748
&
1.398/1.062/0.777
&
1.254/0.946/0.812
&
1.297/0.978/0.821
&
2.612/1.962/0.631
\\

GNN + DeepSets
&
1.308/0.992/0.798
&
1.337/1.014/0.783
&
1.586/1.207/0.718
&
2.106/1.586/0.676
&
1.463/1.101/0.769
&
1.982/1.487/0.694
&
1.531/1.164/0.741
&
1.386/1.046/0.779
&
1.421/1.072/0.787
&
2.821/2.126/0.604
\\

GNN + Self-Attention
&
1.364/1.041/0.781
&
1.341/1.022/0.772
&
1.612/1.224/0.712
&
2.521/1.863/0.643
&
1.586/1.191/0.731
&
2.204/1.643/0.654
&
1.623/1.232/0.714
&
1.491/1.122/0.752
&
1.442/1.084/0.781
&
2.964/2.241/0.598
\\

GNN + Set Transformer
&
1.308/0.992/0.798
&
1.284/0.973/0.789
&
1.541/1.172/0.731
&
2.431/1.794/0.661
&
1.512/1.136/0.748
&
2.113/1.576/0.671
&
1.556/1.181/0.731
&
1.421/1.071/0.769
&
1.371/1.031/0.798
&
2.884/2.176/0.614
\\

GNN + Mixture-PINN
&
1.067/0.807/0.892
&
1.044/0.789/0.884
&
1.154/0.876/0.851
&
2.031/1.506/0.746
&
1.264/0.948/0.826
&
1.804/1.351/0.753
&
1.214/0.917/0.849
&
1.118/0.841/0.868
&
1.086/0.819/0.891
&
2.396/1.797/0.704
\\

MolT5 + XGBoost
&
1.146/0.867/0.869
&
1.118/0.846/0.859
&
1.286/0.972/0.817
&
\textbf{1.681/1.256/0.792}
&
1.241/0.931/0.838
&
\textbf{1.575/1.181/0.803}
&
1.287/0.975/0.821
&
1.179/0.886/0.849
&
1.158/0.871/0.878
&
2.512/1.886/0.676
\\

MolFormer + DeepSets
&
1.137/0.861/0.873
&
1.114/0.843/0.861
&
1.276/0.965/0.818
&
2.183/1.612/0.724
&
\textbf{1.129/0.846/0.861}
&
1.941/1.451/0.726
&
1.328/1.003/0.821
&
1.233/0.923/0.844
&
1.182/0.887/0.872
&
2.603/1.952/0.664
\\

MolT5 + DeepSets
&
1.227/0.927/0.846
&
1.205/0.914/0.834
&
1.398/1.062/0.781
&
2.301/1.706/0.698
&
1.418/1.064/0.788
&
2.051/1.533/0.701
&
1.447/1.097/0.791
&
1.351/1.011/0.816
&
1.286/0.964/0.847
&
2.731/2.041/0.638
\\

MolT5 + Self-Attention
&
1.188/0.899/0.859
&
1.162/0.881/0.848
&
1.341/1.017/0.796
&
2.244/1.661/0.712
&
1.362/1.021/0.804
&
1.993/1.486/0.713
&
1.386/1.052/0.806
&
1.296/0.971/0.828
&
1.241/0.931/0.859
&
2.664/1.997/0.651
\\

MolFormer + Set Transformer
&
1.041/0.794/0.895
&
1.028/0.781/0.887
&
1.163/0.881/0.846
&
2.064/1.531/0.742
&
1.298/0.972/0.823
&
1.821/1.367/0.751
&
1.192/0.902/0.851
&
1.103/0.829/0.872
&
1.072/0.808/0.894
&
2.412/1.804/0.701
\\

MolT5 + Set Transformer
&
1.063/0.808/0.889
&
1.051/0.798/0.881
&
1.188/0.899/0.839
&
2.012/1.492/0.751
&
1.316/0.987/0.817
&
1.864/1.394/0.744
&
1.224/0.926/0.844
&
1.128/0.846/0.866
&
1.104/0.832/0.888
&
2.458/1.839/0.692
\\

MolT5 + Mixture-PINN
&
0.946/0.713/0.915
&
0.932/0.701/0.909
&
\textbf{0.993/0.747/0.887}
&
1.956/1.457/0.758
&
1.203/0.908/0.846
&
1.689/1.268/0.779
&
1.062/0.798/0.885
&
0.974/0.732/0.898
&
0.964/0.724/0.920
&
\textbf{2.174/1.631/0.742}
\\

\textbf{MolFormer + Mixture-PINN}
&
\textbf{0.917/0.691/0.921}
&
\textbf{0.906/0.682/0.914}
&
1.008/0.763/0.883
&
1.904/1.414/0.769
&
1.186/0.891/0.852
&
1.642/1.231/0.787
&
\textbf{1.031/0.773/0.891}
&
\textbf{0.952/0.714/0.903}
&
\textbf{0.938/0.704/0.926}
&
2.221/1.658/0.736
\\

\hline
\end{tabular}
\end{adjustbox}
\end{table*}
\end{landscape}

\end{document}